\documentclass[11pt,a4paper]{article}
\usepackage[hyperref]{acl2019}
\usepackage{times}
\usepackage{latexsym}
\usepackage{url}

\usepackage{graphicx}
\usepackage{wrapfig}
\usepackage{listings}
\usepackage{amsmath}
\usepackage{array,multirow}
\usepackage{makecell}
\usepackage{booktabs, hhline}
\usepackage{color, colortbl}
\usepackage{enumitem}

\aclfinalcopy 

\graphicspath{{figures/}}

\definecolor{light-gray}{gray}{0.9}

\title{ViGGO: A Video Game Corpus for Data-To-Text Generation\\in Open-Domain Conversation}

\author{Juraj Juraska, Kevin K. Bowden \and Marilyn Walker \\
Natural Language and Dialogue Systems Lab \\
University of California, Santa Cruz \\
{\tt \{jjuraska,kkbowden,mawalker\}@ucsc.edu} \\}

\date{}

\begin{document}

\maketitle

\begin{abstract}

The uptake of deep learning in natural language generation (NLG) led to the release of both small and relatively large parallel corpora for training neural models. The existing data-to-text datasets are, however, aimed at task-oriented dialogue systems, and often thus limited in diversity and versatility. They are typically crowdsourced, with much of the noise left in them. Moreover, current neural NLG models do not take full advantage of large training data, and due to their strong generalizing properties produce sentences that look template-like regardless. We therefore present a new corpus of 7K samples, which (1) is clean despite being crowdsourced, (2) has utterances of 9 generalizable and conversational dialogue act types, making it more suitable for open-domain dialogue systems, and (3) explores the domain of video games, which is new to dialogue systems despite having excellent potential for supporting rich conversations.

\end{abstract}

\section{Introduction}
\label{sec:intro}

The recent adoption of deep learning methods in natural language generation (NLG) for dialogue systems resulted in an explosion of neural data-to-text generation models, which depend on large training data. These are typically trained on one of the few parallel corpora publicly available, in particular the E2E~\cite{novikova2017e2e} and the WebNLG~\cite{gardent2017creating} datasets. Crowdsourcing large NLG datasets tends to be a costly and time-consuming process, making it impractical outside of task-oriented dialogue systems. At the same time, current neural NLG models struggle to replicate the high language diversity of the training sentences present in these large datasets, and instead they learn to produce the same generic type of sentences as with considerably less training data~\cite{deriu2018syntactic,juraska2018characterizing,duvsek2019evaluating}.

Motivated by the rising interest in open-domain dialogue systems and conversational agents, we present ViGGO -- a smaller but more comprehensive dataset in the video game domain, introducing several generalizable dialogue acts (DAs), making it more suitable for training versatile and more conversational NLG models.\footnote{The ViGGO corpus is available for download at: https://nlds.soe.ucsc.edu/viggo}
The dataset provides almost 7K pairs of structured meaning representations (MRs) and crowdsourced reference utterances about more than 100 video games. Table~\ref{tab:ex_video_game_dataset} lists three examples.

\begin{table}
    \small
   	\centering
   	\def\arraystretch{1.5}
    \begin{tabular}{p{0.94\linewidth}}
        \hline
    	\rowcolor{light-gray}
    	\emph{give\_opinion}(\textsc{name} [\textbf{SpellForce 3}], \textsc{rating} [\textbf{poor}], \textsc{genres} [\textbf{real-time strategy, role-playing}], \textsc{play\-er\_per\-spec\-tive} [\textbf{bird view}]) \\
        \hline
    	I think that \textbf{SpellForce 3} is \textbf{one of the worst games} I've ever played. Trying to combine the \textbf{real-time strategy} and \textbf{role-playing} genres just doesn't work, and the \textbf{bird's eye view} makes it near impossible to play. \\
        \hline
    	\rowcolor{light-gray}
    	\emph{verify\_attribute}(\textsc{name} [\textbf{Little Big Adventure}], \textsc{rating} [\textbf{average}], \textsc{has\_multiplayer} [\textbf{no}], \textsc{platforms} [\textbf{PlayStation}]) \\
        \hline
    	I recall that you were \textbf{not that fond} of \textbf{Little Big Adventure}. Does \textbf{single-player} gaming on the \textbf{PlayStation} quickly get boring for you? \\
        \hline
    \end{tabular}
	\caption{Examples of MRs and corresponding reference utterances in the ViGGO dataset. The DA of the MRs is indicated in italics, and the slots in small caps. The slot mentions in the utterances are bolded.}
    \label{tab:ex_video_game_dataset}
\end{table}



Video games are a vast entertainment topic that can naturally be discussed in a casual conversation, similar to movies and music, yet in the dialogue systems community it does not enjoy popularity anywhere close to that of the latter two topics \cite{fazel2017learning,li2017end,moghe2018towards,shah2018building,khatri2018advancing}. Restaurants have served as the go-to topic in data-to-text NLG for decades, as they offer a sufficiently large set of various attributes and corresponding values to talk about. While they certainly can be a topic of a casual conversation, the existing restaurant datasets~\cite{stent2004trainable,gavsic2008training,mairesse2010phrase,howcroft2013enhancing,wen2015stochastic,nayak2017plan} are geared more toward a task-oriented dialogue where a system tries to narrow down a restaurant based on the user's preferences and ultimately give a recommendation. Our new video game dataset is designed to be more conversational, and to thus enable neural models to produce utterances more suitable for an open-domain dialogue system.

Even the most recent addition to the publicly available restaurant datasets for data-to-text NLG, the E2E dataset~\cite{novikova2017e2e}, suffers from the lack of a conversational aspect. It has become popular, thanks to its unprecedented size and multiple reference utterances per MR, for training end-to-end neural models, yet it only provides a single DA type. In contrast with the E2E dataset, ViGGO presents utterances of 9 different DAs.

Other domains have been represented by task-oriented datasets with multiple DA types, for example the Hotel, Laptop, and TV datasets~\cite{wen2015semantically,wen2016multi}. Nevertheless, the DAs in these datasets vary greatly in complexity, and their distribution is thus heavily skewed, typically with two or three similar DAs comprising almost the entire dataset. In our video game dataset, we omitted simple DAs, in particular those that do not require any slots, such as greetings or short prompts, and focused on a set of substantial DAs only.

The main contribution of our work is thus a new parallel data-to-text NLG corpus that (1) is more conversational, rather than information seeking or question answering, and thus more suitable for an open-domain dialogue system, (2) represents a new, unexplored domain which, however, has excellent potential for application in conversational agents, and (3) has high-quality, manually cleaned human-produced utterances.

\section{The ViGGO Dataset}

ViGGO features more than 100 different video game titles, whose attributes were harvested using free API access to two of the largest online video game databases: IGDB\footnote{https://www.igdb.com/} and GiantBomb\footnote{https://www.giantbomb.com/}. Using these attributes, we generated a set of 2,300 structured MRs. The human reference utterances for the generated MRs were then crowdsourced using vetted workers on the Amazon Mechanical Turk (MTurk) platform~\cite{buhrmester2011amazon}, resulting in 6,900 MR-utterance pairs altogether. With the goal of creating a clean, high-quality dataset, we strived to obtain reference utterances with correct mentions of all slots in the corresponding MR through post-processing.

\subsection{Meaning Representations}
\label{sec:meaning_representations}

\begin{figure}
    \begin{center}
        \includegraphics[width=\columnwidth]{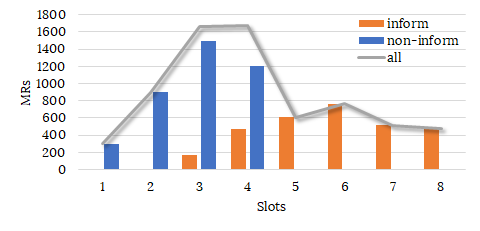}
    \end{center}
    \vspace{-0.7cm}
    \caption{Distribution of the number of slots across all types of MRs, as well as the \emph{inform} slot separately, and non-\emph{inform} slots only.}
    \label{fig:slot_per_mr_distr}
    \vspace{-0.3cm}
\end{figure}

The MRs in the ViGGO dataset range from 1 to 8 slot-value pairs, and the slots come from a set of 14 different video game attributes. Table~\ref{tab:video_game_da_slots} details how these slots may be distributed across the 9 different DAs. The \emph{inform} DA, represented by 3,000 samples, is the most prevalent one, as the average number of slots it contains is significantly higher than that of all the other DAs. Figure~\ref{fig:slot_per_mr_distr} visualizes the MR length distribution across the entire dataset.

\begin{table}
    \small
   	\centering
    \begin{tabular}{| >{\centering\arraybackslash} m{0.18\linewidth} | >{\centering\arraybackslash} m{0.08\linewidth} | >{\centering\arraybackslash} m{0.18\linewidth} | >{\centering\arraybackslash} m{0.33\linewidth} |}
    	\hline
    	\textbf{DA} & \textbf{Slot range}   & \textbf{Mandatory slots}    & \textbf{Additional common slots} \\
        \hline
    	\emph{inform}   & 3-8   & \textsc{name, genres} & \multirow{9}{\linewidth}{\centering \textsc{release\_year, developer, esrb, genres, play\-er\_per\-spec\-tive, has\_multi\-player, plat\-forms, avail\-able\_on\_steam, has\_linux\_re\-lease, has\_mac\_release}} \\
        \cline{1-3}
    	\emph{confirm}   & 2-3   & \textsc{name}    & \\
        \cline{1-3}
    	\emph{give\_opin\-ion}   & 3-4   & \textsc{name, rating} & \\
        \cline{1-3}
    	\emph{recommend}   & 2-3    & \textsc{name} & \\
        \cline{1-3}
    	\emph{request}   & 1-2  & \textsc{specifier}    & \\
        \cline{1-3}
    	\emph{request\_at\-tribute}   & 1  & & \\
        \cline{1-3}
    	\emph{request\_ex\-planation}   & 2-3   & \textsc{rating} & \\
        \cline{1-3}
    	\emph{suggest}   & 2-3  & \textsc{name} & \\
        \cline{1-3}
    	\emph{verify\_at\-tribute}   & 3-4  & \textsc{name, rating} & \\
        \hline
    \end{tabular}
	\caption{Overview of mandatory and common possible slots for each DA in the ViGGO dataset. There is an additional slot, \textsc{exp\_re\-lease\_date}, only possible in the \emph{inform} and \emph{confirm} DAs. Moreover, \textsc{rating} is also possible in the \emph{inform} DA, though not mandatory.}
    \label{tab:video_game_da_slots}
\end{table}

The slots can be classified into 5 general categories covering most types of information MRs typically convey in data-to-text generation scenarios:
\emph{Boolean}, \emph{Numeric}, \emph{Scalar}, \emph{Categorical}, and \emph{List}.
The first 4 categories are common in other NLG datasets, such as E2E, Laptop, TV, and Hotel, while the \emph{List} slots are unique to ViGGO. \emph{List} slots have values which may comprise multiple items from a discrete list of possible items.

\subsection{Utterances}

With neural language generation in mind, we crowdsourced 3 reference utterances for each MR so as to provide the models with the information about how the same content can be realized in multiple different ways. At the same time, this allows for a more reliable automatic evaluation by comparing the generated utterances with a set of different references each, covering a broader spectrum of correct ways of expressing the content given by the MR. The raw data, however, contains a significant amount of noise, as is inevitable when crowdsourcing. We therefore created and enforced a robust set of heuristics and regular expressions to account for typos, grammatical errors, undesirable abbreviations, unsolicited information, and missing or incorrect slot realizations.

\subsection{Data Collection}

The crowdsourcing of utterances on MTurk took place in three stages. After collecting one third of the utterances, we identified a pool of almost 30 workers who wrote the most diverse and natural-sounding sentences in the context of video games. We then filtered out all utterances of poor quality and had the qualified workers write new ones for the corresponding inputs. Finally, the remaining two thirds of utterances were completed by these workers exclusively.

For each DA we created a separate task in order to minimize the workers' confusion. The instructions contained several different examples, as well as counter-examples, and they situated the DA in the context of a hypothetical conversation. The video game attributes to be used were provided for the workers in the form of a table, with their order shuffled so as to avoid any kind of bias. Further details on the data collection and cleaning are included in the Appendix.

\subsection{Train/Validation/Test Split}

\begin{figure}
    \begin{center}
        \includegraphics[width=\columnwidth]{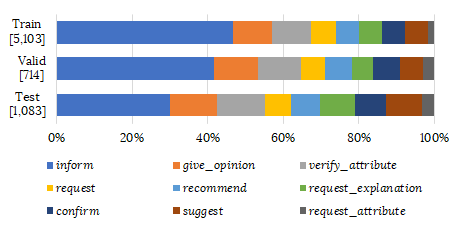}
    \end{center}
    \vspace{-0.2cm}
    \caption{Distribution of the DAs across the train/validation/test split. For each partition the total count of DAs/MRs is indicated.}
    \label{fig:da_per_train_valid_test_distr}
\end{figure}

Despite the fact that the ViGGO dataset is not very large, we strived to make the test set reasonably challenging. To this end, we ensured that, after delexicalizing the \textsc{name} and the \textsc{developer} slots, there were no common MRs between the train set and either of the validation or test set. We maintained a similar MR length and slot distribution across the three partitions. The distribution of DA types, on the other hand, is skewed slightly toward fewer \emph{inform} DA instances and a higher proportion of the less prevalent DAs in the validation and test sets (see Figure~\ref{fig:da_per_train_valid_test_distr}). With the exact partition sizes indicated in the diagram, the final ratio of samples is approximately $7.5:1:1.5$.

\subsection{ViGGO vs. E2E}
\label{sec:vg_vs_e2e}

Our new dataset was constructed under different constraints than the E2E dataset. First, in ViGGO we did \emph{not} allow any omissions of slot mentions, as those are not justifiable for data-to-text generation with no previous context, and it makes the evaluation ambiguous. Second, the MRs in ViGGO are grounded by \emph{real} video game data, which can encourage richer and more natural-sounding reference utterances.

While ViGGO is only 13\% the size of the E2E dataset, the lexical diversity of its utterances is 77\% of that in the E2E dataset, as indicated by the ``delexicalized vocabulary'' column in Table~\ref{tab:vg_vs_e2e_dataset_stats}. Part of the reason naturally is the presence of additional DAs in ViGGO, and therefore we also indicate the statistics in Table~\ref{tab:vg_vs_e2e_dataset_stats} for the \emph{inform} samples only. The average \emph{inform} utterance length in ViGGO turns out to be over 30\% greater, in terms of both words and sentences per utterance.

Finally, we note that, unlike the E2E dataset, our test set does not place any specific emphasis on longer MRs. While the average number of slots per MR in the \emph{inform} DAs are comparable to the E2E dataset, in general the video game MRs are significantly shorter. This is by design, as shorter, more focused responses are more conversational than consistently dense utterances.

\begin{table*}
    \small
   	\centering
    \begin{tabular}{>{\centering\arraybackslash} m{0.09\linewidth} >{\centering\arraybackslash} m{0.07\linewidth} >{\centering\arraybackslash} m{0.06\linewidth} >{\centering\arraybackslash} m{0.06\linewidth} >{\centering\arraybackslash} m{0.05\linewidth} >{\centering\arraybackslash} m{0.05\linewidth} >{\centering\arraybackslash} m{0.06\linewidth} >{\centering\arraybackslash} m{0.04\linewidth} >{\centering\arraybackslash} m{0.05\linewidth} >{\centering\arraybackslash} m{0.05\linewidth} >{\centering\arraybackslash} m{0.05\linewidth} >{\centering\arraybackslash} m{0.05\linewidth}}
    	\toprule
    	& \textbf{Instances}    & \textbf{Unique MRs}   & \textbf{Unique delex. MRs}  & \textbf{Vocab}    & \textbf{Delex. vocab}   & \textbf{Avg. 3-gram freq.}  & \textbf{Refs/ MR}    & \textbf{Slots/ MR}   & \textbf{W/ Ref} & \textbf{W/ Sent}   & \textbf{Sents/ Ref} \\
    	\midrule
    	\textbf{E2E}    & 51,426    & 6,039 & 5,963  & 2,878    & 2,818   & 18.70   & 8.1   & 5.43  & 22.41 & 14.36  & 1.56 \\
    	\textbf{ViGGO\textsubscript{\emph{inf}}}    & 3,000    & 1,000 & 997 & 1,378 & 1,102 & 8.33 & 3   & 5.81  & 30.62 & 15.01  & 2.04 \\
    	\textbf{ViGGO}    & 6,900    & 2,253 & 2,066   & 2,427 & 2,178   & 6.91   & 3   & 4.18  & 25.01 & 15.04  & 1.66 \\
        \bottomrule
    \end{tabular}
	\caption{Dataset statistics comparing the ViGGO dataset, as well as its subset of \emph{inform} DAs only (ViGGO\textsubscript{\emph{inf}}), with the E2E dataset. The average trigram frequency was calculated on trigrams that appear more than once.}
    \label{tab:vg_vs_e2e_dataset_stats}
\end{table*}

\section{Baseline System Evaluation}
\label{sec:evaluation}

The NLG model we use to establish a baseline for this dataset is a standard Transformer-based~\cite{vaswani2017attention} sequence-to-sequence model. For decoding we employ beam search of width 10 ($\alpha = 1.0$). The generated candidates are then reranked according to the heuristically determined slot coverage score. Before training the model on the ViGGO dataset, we confirmed on the E2E dataset that it performed on par with, or even slightly better than, the strong baseline models from the E2E NLG Challenge\footnote{http://www.macs.hw.ac.uk/InteractionLab/E2E/}, namely, \textsc{TGen}~\cite{duvsek2016sequence} and \textsc{Slug2Slug}~\cite{juraska2018deep}.

\paragraph{Automatic Metrics}
We evaluate our model's performance on the ViGGO dataset using the following standard NLG metrics: BLEU~\cite{papineni2002bleu}, METEOR~\cite{lavie2007meteor}, ROUGE-L~\cite{lin2004rouge}, and CIDEr~\cite{vedantam2015cider}. Additionally, with our heuristic slot error rate (SER) metric we approximate the percentage of failed slot realizations (i.e., missed, incorrect, or hallucinated) across the test set. The results are shown in Table~\ref{tab:results_automatic_metrics}.

\begin{table}
    \small
    \centering
        \begin{tabular} { >{\centering\arraybackslash}m{0.5cm} >{\centering\arraybackslash}m{0.8cm} >{\centering\arraybackslash}m{1.2cm} >{\centering\arraybackslash}m{1.0cm} >{\centering\arraybackslash}m{0.9cm} >{\centering\arraybackslash}m{0.7cm}
        }
        \toprule
        & \textbf{BLEU}
        & \textbf{METEOR}
        & \textbf{ROUGE}
        & \textbf{CIDEr}
        & \textbf{SER} \\
        \midrule
        \textbf{Ao3}	& 0.519   & 0.388  & 0.631  & 2.531  & \emph{2.55\%} \\
    	\textcolor{gray}{\textbf{Bo3}}		& \textcolor{gray}{0.521}   & \textcolor{gray}{0.391}  & \textcolor{gray}{0.638}  & \textcolor{gray}{2.545}  & \textcolor{gray}{\emph{2.48\%}} \\
        \bottomrule
    \end{tabular}
    \caption{Baseline system performance on the ViGGO test set. Despite individual models (Bo3 -- best of 3 experiments) often having better overall scores, we consider the Ao3 (average of 3) results the most objective.}
    \label{tab:results_automatic_metrics}
\end{table}

\paragraph{Human Evaluation}
We let two expert annotators with no prior knowledge of the ViGGO dataset evaluate the outputs of our model. Their task was to rate 240 shuffled utterances (120 generated utterances and 120 human references) each on \emph{naturalness} and \emph{coherence} using a 5-point Likert scale. We define naturalness as a measure of how much one would expect to encounter an utterance in a conversation with a human, as opposed to sounding robotic, while coherence measures its grammaticality and fluency. Out of the 120 MRs in each partition, 40 were of the \emph{inform} type, with the other 8 DAs represented by 10 samples each. In addition to that, we had the annotators rate a sample of 80 utterances from the E2E dataset (40 generated and 40 references) as a sort of a baseline for the human evaluation.

With both datasets, our model's outputs were highly rated on both naturalness and coherence (see Table~\ref{tab:results_human_eval}). The scores for the ViGGO utterances were overall higher than those for the E2E ones, which we understand as an indication of the video game data being more fluent and conversational. At the same time, we observed that the utterances generated by our model tended to score higher than the reference utterances, though significantly more so for the E2E dataset. This is likely a consequence of the ViGGO dataset being cleaner and less noisy than the E2E dataset.

\begin{table}
    \small
   	\centering
    \begin{tabular}{>{\centering\arraybackslash} m{0.18\linewidth} >{\centering\arraybackslash} m{0.11\linewidth} >{\centering\arraybackslash} m{0.16\linewidth} >{\centering\arraybackslash} m{0.11\linewidth} >{\centering\arraybackslash} m{0.16\linewidth}}
    	\toprule
    	& \multicolumn{2}{c}{\textbf{Naturalness}}    & \multicolumn{2}{c}{\textbf{Coherence}} \\
    	& \textbf{Ref.} & \textbf{Gen. utt.}    & \textbf{Ref.} & \textbf{Gen. utt.} \\
    	\midrule
    	\textbf{E2E}    & 4.48  & 4.67  & 4.57  & 4.77 \\
    	\textbf{ViGGO\textsubscript{\emph{inf}}}    & 4.85  & 4.83  & 4.85  & 4.93 \\
    	\textbf{ViGGO}    & 4.68  & 4.74  & 4.78  & 4.84 \\
        \bottomrule
    \end{tabular}
	\caption{Naturalness and coherence scores of our model's generated outputs compared to the reference utterances, as per the human evaluation. ViGGO\textsubscript{\emph{inf}} corresponds to the subset of \emph{inform} DAs only.}
    \label{tab:results_human_eval}
\end{table}

In an additional evaluation of ViGGO, we asked the annotators to classify the utterance samples into the 9 DA groups. For this task they were provided with a brief description of each DA type. The annotators identified the DA incorrectly in only 7\% of the samples, which we interpret as a confirmation that our DAs are well-defined. Most of the mistakes can be ascribed to the inherent similarity of the \emph{recommend} and the \emph{suggest} DA, as well as to our model often generating \emph{give\_opinion} utterances that resemble the \emph{inform} ones.

\paragraph{Qualitative Analysis}
Among all 9 DAs, the one posing the greatest challenge for our model was \emph{give\_opinion}, due to its high diversity of reference utterances. Despite the occasional incoherence, it learned to produce rich and sensible utterances, for instance ``Little Nightmares is a pretty good game. Tarsier Studios is a talented developer and the side view perspective makes it easy to play.''.

Since our baseline model does not implement any form of a copy mechanism, it fails on instances with out-of-vocabulary terms, such as the values of the \textsc{specifier} slot in the test set. These, in fact, account for almost half of the errors indicated by the SER metric in Table~\ref{tab:results_automatic_metrics}. Therefore, more robust models have good potential for improving on our scores.

\section{Discussion}

In Table~\ref{tab:ex_video_game_conversation} we demonstrate how the 9 DAs of the ViGGO dataset can support a natural multi-turn exchange on the topic of video games, as a part of a longer casual conversation on different topics. One caveat of using a language generator trained on this dataset in a dialogue system as-is is that multiple subsequent turns discussing the same video game would be repeating its full name. ViGGO was designed for grounded generation but without context, and therefore it is up to the dialogue manager to ensure that pronouns are substituted for the names whenever it would sound more natural in a dialogue. Alternately, the dataset can easily be augmented with automatically constructed samples which omit the \textsc{name} slot in the MR and replace the name with a pronoun in the reference utterance.

\begin{table}
    \small
   	\centering
   	\def\arraystretch{1.5}
    \begin{tabular}{p{0.02\linewidth} >{\raggedright\arraybackslash}p{0.87\linewidth}}
		\hline
    	S:  & \dots What about video games? \\
    	\hline
    	\rowcolor{light-gray}
    	U:  & Totally, I spend my entire evenings in front of my PS4. \\
    	\hline
    	S:  & What is your favorite \textbf{genre} of video games? \\
    	& \textcolor{darkgray}{\emph{request\_attribute}(\textsc{genres} [])} \\
    	\hline
    	\rowcolor{light-gray}
    	U:  & I guess I play a lot of shooter games. \\
    	\hline
    	S:  & Have you played any \textbf{interesting} games lately? \\
    	& \textcolor{darkgray}{\emph{request}(\textsc{specifier} [\textbf{interesting}])} \\
    	\hline
    	\rowcolor{light-gray}
    	U:  & Tomb Raider is pretty good. \\
    	\hline
    	S:  & Do you mean \textbf{Tomb Raider} from \textbf{2013}? \\
    	& \textcolor{darkgray}{\emph{confirm}(\textsc{name} [\textbf{Tomb Raider}], \textsc{release\_year} [\textbf{2013}])} \\
    	\hline
    	\rowcolor{light-gray}
    	U:  & Yes, that one with the epic E3 trailer. \\
    	\hline
    	S:  & Do you also enjoy playing \textbf{first-person} \textbf{shooters}, such as \textbf{Half-Life 2}? \\
    	& \textcolor{darkgray}{\emph{suggest}(\textsc{name} [\textbf{Half-Life 2}], \textsc{genres} [\textbf{shooter}], \textsc{player\_perspective} [\textbf{first person}])} \\
    	\hline
    	\rowcolor{light-gray}
    	U:  & I haven't played that one in a while. Would you recommend something else? \\
    	\hline
    	S:  & Well, since you seem to be a fan of \textbf{M-rated} \textbf{PlayStation} games, I thought you would perhaps like \textbf{Max Payne 3}. Have you heard about that game? \\
    	& \textcolor{darkgray}{\emph{recommend}(\textsc{name} [\textbf{Max Payne 3}], \textsc{esrb} [\textbf{M (for Mature)}], \textsc{platforms} [\textbf{PlayStation}])} \\
    	\hline
    	\rowcolor{light-gray}
    	U:  & No. \\
    	\hline
    	S:  & \textbf{Max Payne 3} is a \textbf{multiplayer} \textbf{action-adventure} \textbf{shooter} from \textbf{2012}. \\
    	& \textcolor{darkgray}{\emph{inform}(\textsc{name} [\textbf{Max Payne 3}], \textsc{release\_year} [\textbf{2012}], \textsc{genres} [\textbf{action-adventure; shooter}], \textsc{multiplayer} [\textbf{yes}])} \\
    	\hline
    	\rowcolor{light-gray}
    	U:  & I see. So, you enjoyed it? \\
    	\hline
    	S:  & Oh, I \textbf{love} \textbf{Max Payne 3}. When I see a game developed by \textbf{Rockstar Games}, I buy it without thinking. \\
    	& \textcolor{darkgray}{\emph{give\_opinion}(\textsc{name} [\textbf{Max Payne}], \textsc{rating} [\textbf{excellent}], \textsc{developer} [\textbf{Rockstar Games}])} \\
    	\hline
    	\rowcolor{light-gray}
    	U:  & That sounds cool. \\
    	\hline
    	S:  & Previously you told me that you \textbf{enjoyed} \textbf{Tomb Raider}. Is that the case with all games developed by \textbf{Crystal Dynamics}? \\
    	& \textcolor{darkgray}{\emph{verify\_attribute}(\textsc{name} [\textbf{Tomb Raider}], \textsc{rating} [\textbf{good}], \textsc{developer} [\textbf{Crystal Dynamics}])} \\
    	\hline
    	\rowcolor{light-gray}
    	U:  & You bet. \\
    	\hline
    	S:  & What is it that you \textbf{like} about \textbf{Crystal Dynamics} games? \\
    	& \textcolor{darkgray}{\emph{request\_explanation}(\textsc{rating} [\textbf{good}], \textsc{developer} [\textbf{Crystal Dynamics}])} \\
    	\hline
    	\rowcolor{light-gray}
    	U:  & They usually have really good stories. \\
    	\hline
    	S:  & \dots \\
    	\hline
    \end{tabular}
 	\vspace{-0.1cm}
	\caption{An example of a chit-chat about video games comprising utterances of DAs defined in ViGGO. ``S''~denotes the system and ``U'' the user turns.}
	\vspace{-0.4cm}
    \label{tab:ex_video_game_conversation}
 	\vspace{-0.1cm}
\end{table}

\section{Conclusion}
\label{sec:conclusion}

In this paper we presented a new parallel corpus for data-to-text NLG, which contains 9 dialogue acts, making it more conversational than other similar datasets. The crowdsourced utterances were thoroughly cleaned in order to obtain high-quality human references, which we hope will support the recent trend in research to train neural models on small but high-quality data, like humans can. This could possibly be achieved by transferring fundamental knowledge from larger available corpora, such as the E2E dataset, but perhaps by other, completely new, methods.

\bibliography{references}
\bibliographystyle{acl_natbib}


\appendix

\section{Appendix}
\label{sec:appendix}

\subsection{Additional ViGGO Dataset Examples}

In Table~\ref{tab:ex_video_game_dataset_full} we present one example of each DA in the ViGGO dataset, including the examples given in Table~\ref{tab:ex_video_game_dataset}.

\begin{table}
    \small
  	\centering
  	\def\arraystretch{1.5}
    \begin{tabular}{p{0.94\linewidth}}
    	\hline
    	\rowcolor{light-gray}
    	\emph{inform}(\textsc{name} [\textbf{God of War}], \textsc{release\_year} [\textbf{2018}], \textsc{developer} [\textbf{SIE Santa Monica Studio}], \textsc{rating} [\textbf{excellent}], \textsc{genres} [\textbf{action-adventure, platformer, role-playing}], \textsc{player\_perspective} [\textbf{third person}], \textsc{has\_multiplayer} [\textbf{no}], \textsc{platforms} [\textbf{PlayStation}]) \\
        \hline
    	Developed by \textbf{SIE Santa Monica Studio} in \textbf{2018}, \textbf{God of War} is an \textbf{excellent} \textbf{single-player} \textbf{third person} \textbf{platformer} made exclusively for \textbf{PlayStation}. The \textbf{action-adventure} storyline involves \textbf{role-playing} as one of the dynamic characters. \\
    	\hline
    	\rowcolor{light-gray}
    	\emph{confirm}(\textsc{name} [\textbf{Hellblade: Senua's Sacrifice}], \textsc{release\_year} [\textbf{2017}], \textsc{developer} [\textbf{Ninja Theory}]) \\
        \hline
    	Oh, do you mean the \textbf{2017} game from \textbf{Ninja Theory}, \textbf{Hellblade: Senua's Sacrifice}? \\
        \hline
    	\rowcolor{light-gray}
    	\emph{give\_opinion}(\textsc{name} [\textbf{SpellForce 3}], \textsc{rating} [\textbf{poor}], \textsc{genres} [\textbf{real-time strategy, role-playing}], \textsc{play\-er\_per\-spec\-tive} [\textbf{bird view}]) \\
        \hline
    	I think that \textbf{SpellForce 3} is \textbf{one of the worst games} I've ever played. Trying to combine the \textbf{real-time strategy} and \textbf{role-playing} genres just doesn't work, and the \textbf{bird's eye view} makes it near impossible to play. \\
        \hline
    	\rowcolor{light-gray}
    	\emph{recommend}(\textsc{name} [\textbf{Call of Duty: Advanced Warfare}], \textsc{developer} [\textbf{Sledgehammer Games}], \textsc{esrb} [\textbf{M (for Mature)}]) \\
        \hline
    	Speaking of \textbf{M rated} games developed by \textbf{Sledgehammer Games}, have you tried \textbf{Call of Duty: Advanced Warfare}? \\
        \hline
    	\rowcolor{light-gray}
    	\emph{request}(\textsc{developer} [\textbf{Guerrilla Games}], \textsc{specifier} [\textbf{overrated}]) \\
        \hline
    	What would you say is the most \textbf{overrated} game made by \textbf{Guerrilla Games}? \\
        \hline
    	\rowcolor{light-gray}
    	\emph{request\_attribute}(\textsc{available\_on\_steam} []) \\
        \hline
    	Do you prefer playing games that you can get on \textbf{Steam}? \\
        \hline
    	\rowcolor{light-gray}
    	\emph{request\_explanation}(\textsc{rating} [\textbf{poor}], \textsc{\textbf{has\_mac\_re\-lease}} [yes]) \\
        \hline
    	What is it about \textbf{Mac} games that you find \textbf{so disappointing}? \\
        \hline
    	\rowcolor{light-gray}
    	\emph{suggest}(\textsc{name} [\textbf{Rocket League}], \textsc{genres} [\textbf{sport, vehicular combat}], \textsc{player\_perspective} [\textbf{third person}]) \\
        \hline
    	Are you into \textbf{third person} \textbf{sport} games with \textbf{vehicular combat} like \textbf{Rocket League}? \\
        \hline
    	\rowcolor{light-gray}
    	\emph{verify\_attribute}(\textsc{name} [\textbf{Little Big Adventure}], \textsc{rating} [\textbf{average}], \textsc{has\_multiplayer} [\textbf{no}], \textsc{platforms} [\textbf{PlayStation}]) \\
        \hline
    	I recall that you were \textbf{not that fond} of \textbf{Little Big Adventure}. Does \textbf{single-player} gaming on the \textbf{PlayStation} quickly get boring for you? \\
        \hline
    \end{tabular}
	\caption{Examples of MRs and corresponding reference utterances in the ViGGO dataset. The DA of the MRs is indicated in italics, and the slots in small caps. The slot mentions in the utterances are bolded.}
    \label{tab:ex_video_game_dataset_full}
\end{table}

\subsection{Slot Categories}

In Section~\ref{sec:meaning_representations} we mentioned that the slots in the ViGGO dataset can be classified into 5 general categories. Here we provide more detailed descriptions of the categories:
\begin{enumerate}[itemsep=0pt, parsep=0pt]
    \item \emph{Boolean} -- binary value, such as ``yes''/``no'' or ``true''/``false'' (e.g., \textsc{has\_multiplayer} or \textsc{available\_on\_steam}),
    \item \emph{Numeric} -- value is a number or contains number(s) as the salient part (e.g., \textsc{release\_year} or \textsc{exp\_release\_date}),
    \item \emph{Scalar} -- values are on a distinct scale (e.g., \textsc{rating} or \textsc{esrb}),
    \item \emph{Categorical} -- takes on virtually any value, typically coming from a certain category, such as names or types (e.g., \textsc{name} or \textsc{developer}),
    \item \emph{List} -- similar to categorical, where the value can, however, consist of multiple individual items (e.g., \textsc{genres} or \textsc{player\_perspective}).
\end{enumerate}
Note that in ViGGO the items in the value of a \emph{List} slot are comma-separated, and therefore the individual items must not contain a comma. There are no restrictions as to whether the values are single-word or multi-word in any of the categories.

\subsection{Data Collection}




When generating the MRs for the \emph{inform} DA, we fixed the slot ratios: the \textsc{name} and \textsc{genres} slots were mandatory in every MR, the \textsc{player\_perspective} and \textsc{release\_year} were enforced in about half of the MRs, while the remaining slots are present in about 25\% of the MRs. At the same time we imposed two constraints on the slot combinations: (1)~whenever one of the Steam, Linux or Mac related boolean slots is present in an MR, the \textsc{platforms} slot must be included too, and (2)~whenever either of the Linux or Mac slots was picked for an MR, the other one was automatically added too. These two constraints were introduced so as to encourage reference utterances with natural aggregations and contrast relations.

The remaining 8 DAs, however, contain significantly fewer slots each (see Table~\ref{tab:video_game_da_slots}). We therefore decided to have the MTurk workers select 5 unique slot combinations for each given video game before writing the corresponding utterances. Since for these DAs we collected less data, we tried to ensure in this way that we have a sufficient number of samples for those slot combinations that are most natural to be mentioned in each of the DAs. While fixing mandatory slots for each DA, we instructed the workers to choose 1 or 2 additional slots depending on the task. The data collection for MRs with only 1 additional slot and for those with 2 was performed separately, so as to prevent workers from taking the easy way out by always selecting just a single slot, given the option.

\begin{figure}
    \begin{center}
        \includegraphics[width=\columnwidth]{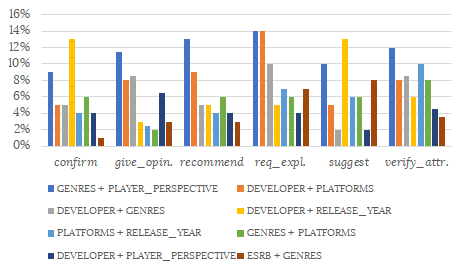}
    \end{center}
    \caption{Distribution of the 8 most frequently selected slot combinations across different DAs.}
    \label{fig:selected_slot_combo_distr_2slots_top8}
\end{figure}

Leaving the slot selection to crowdworkers yields a frequency distribution of all slot combinations, which presumably indicates the suitability of different slots to be mentioned together in a sentence. This meta-information can be made use of in a system's dialogue manager to sample from the observed slot combination distributions instead of sampling randomly or hard-coding the combinations. Figure~\ref{fig:selected_slot_combo_distr_2slots_top8} shows the distributions of the 8 slot pairs most commonly mentioned together in different DAs. These account for 53\% of the selections among the 6 DAs that can take 2 additional slots besides the mandatory ones. We can observe some interesting trends in the distributions, such as that the \textsc{developer~+~release\_year} combination was the most frequent one in the \emph{confirm} DA, while fairly rare in most of the other DAs. This might be because this pair of a game's attributes is arguably the next best identifier of a game after its name.

\subsection{Dataset Cleaning}

A large proportion of the raw data collected contained typos and various errors, as is inevitable when crowdsourcing. We took the following three steps to clean the data.

First, we used regular expressions to enforce several standardization policies regarding special characters, punctuation, and the correction of undesired abbreviations/misspellings of standard domain-specific terms (e.g., we would change terms like ``Play station'' or ``PS4'' to the uniform ``PlayStation''). At the same time, we removed or enforced hyphens uniformly in certain terms, for example, ``single-player''. Although phrases such as ``first person'' should correctly have a hyphen when used as adjective, the turkers used this rule very inconsistently. In order to avoid model outputs being penalized during the evaluation by the arbitrary choice of a hyphen presence or absence in the reference utterances, we decided to remove the hyphen in all such phrases regardless of the noun/adjective use.

Second, we developed an extensive set of heuristics to identify slot-related errors. This process revealed the vast majority of missing or incorrect slot mentions, which we subsequently fixed according to the corresponding MRs. Turkers would sometimes also inject a piece of information which was not present in the MR, some of which is not even represented by any of the slots, e.g., plot or main characters. We remove this extraneous information from the utterances so as to avoid confusing the neural model. This step thus involved certain manual work and was thus performed jointly with the third step.

Finally, we further resolved the remaining typos, grammatical errors, and unsolicited information.

\subsection{Model Parameters}

Even though on the small datasets we work with we do not necessarily expect the Transformer model to perform better than recurrent neural networks, we chose this model for its significantly faster training, without sacrificing the performance. For our experiments a small 2-layer Transformer with 8 heads proved to be sufficient. The input tokens are encoded into embeddings of size 256, and the target sequences were truncated to 60 tokens. The model performed best with dropout values of 0.2. For training of the Transformer models we used the Adam optimizer with a custom learning rate schedule including a brief linear warm-up and a cosine decay.


\end{document}